\documentclass[11pt]{article}

\usepackage[preprint]{acl}

\usepackage{times}
\usepackage{latexsym}
\usepackage[T1]{fontenc}
\usepackage[utf8]{inputenc}
\usepackage{microtype}
\usepackage{inconsolata}

\usepackage{graphicx}
\usepackage{array}
\usepackage{siunitx}
\usepackage{booktabs}
\usepackage{caption}
\usepackage{comment}
\usepackage{placeins}
\usepackage{subcaption}
\usepackage{amsmath}

\usepackage{listings} 

\usepackage{color}
\definecolor{codegreen}{rgb}{0,0.6,0}
\definecolor{codegray}{rgb}{0.5,0.5,0.5}
\definecolor{codepurple}{rgb}{0.58,0,0.82}
\definecolor{backcolour}{rgb}{0.95,0.95,0.95}

\newcommand{\ra}[1]{\renewcommand{\arraystretch}{#1}}

\title{Cascaded Information Disclosure for Generalized Evaluation of Problem Solving Capabilities}

\author{Yunxiang Yan \quad Tomohiro Sawada \quad Kartik Goyal \\
  College of Computing \\
  Georgia Institute of Technology \\
  \texttt{\{ryan.yunxiang.yan, tsawada, kartikgo\}@gatech.edu}
}

\begin{document}
\maketitle

\begin{abstract}
While question-answering~(QA) benchmark performance is an automatic and scalable method to compare LLMs, it is an indirect method of evaluating their underlying problem-solving capabilities.
Therefore, we propose a holistic and generalizable framework based on \emph{cascaded question disclosure} that provides a more accurate estimate of the models' problem-solving capabilities while maintaining the scalability and automation. 
This approach collects model responses in a stagewise manner with each stage revealing partial information about the question designed to elicit generalized reasoning in LLMs. 
We find that our approach not only provides a better comparison between LLMs, but also induces better intermediate traces in models compared to the standard QA paradigm. 
We empirically verify this behavior on diverse reasoning and knowledge-heavy QA datasets by comparing LLMs of varying sizes and families. 
Our approach narrows the performance gap observed in the standard QA evaluation settings, indicating that the prevalent indirect QA paradigm of evaluation overestimates the differences in performance between models.
We further validate our findings by extensive ablation studies. 

\end{abstract}
\begin{figure*}[t] 
    \centering
    \includegraphics[width=\textwidth]{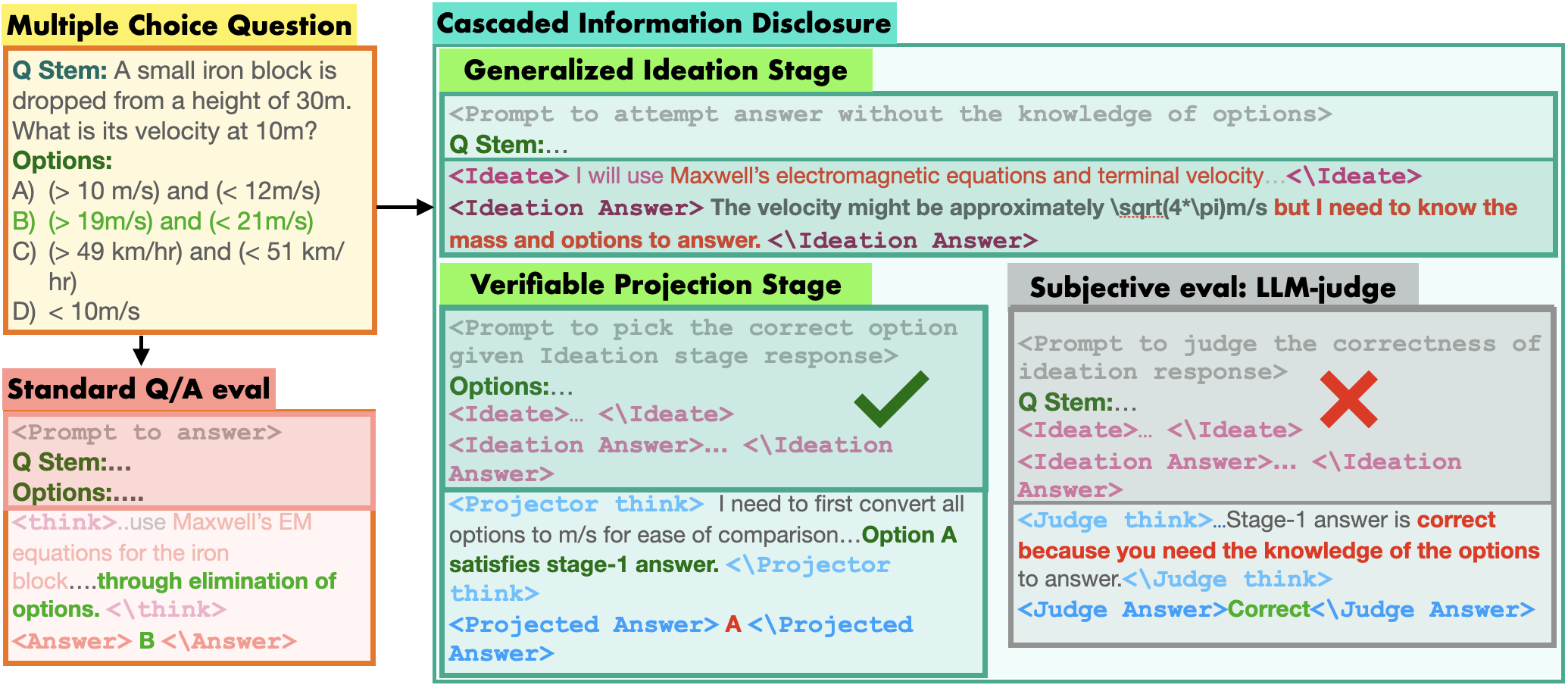} 
    \caption{Demonstration of MCQA instantiation of our Cascaded Information Disclosure evaluation framework against Standard evaluation and LLM-based Subjective evaluation of problem-solving capabilities. The \textbf{standard evaluation} paradigm picks the correct option but has poor justification for the answer. Cascaded Information Disclosure strategically generalizes the question and exposes parts of it to the model in a stagewise manner. The free-form response generated during \textbf{ideation} phase reflects model's poor problem-solving ability. \textbf{LLM-based subjective evaluation} is an appealing alternative to judge the ideated response but fails in the example shown -- it supports the trivial conclusion in the incorrect ideation. \textbf{Verifiable Projection} while objective and automatic, correctly penalizes the incorrect ideated response by picking the wrong option.}
    \label{fig:1_wide} 
\end{figure*}
\section{Introduction}
While general-purpose LLMs have become ubiquitous today, evaluating them holistically remains a massive challenge. Primarily, these models are compared against each other by their performance on a handful of benchmark tasks and datasets~\cite{liang2023holisticevaluationlanguagemodels,srivastava2023imitationgamequantifyingextrapolating,wang2019gluemultitaskbenchmarkanalysis,sarlin2020supergluelearningfeaturematching} dominated by various kinds of question-answering tasks~\cite{hendrycks2021measuringmassivemultitasklanguage,lin2022truthfulqameasuringmodelsmimic,rajpurkar2016squad100000questionsmachine,rein2023gpqagraduatelevelgoogleproofqa}. While objective and scalable, to estimate the underlying problem-solving capabilities of models, this approach is \emph{indirect} -- merely judging a model on its ability to pick the correct choice among distractors or to produce a correct numerical answer to a mathematical problem doesn’t measure the quality of strategies these models use to arrive at that answer. There is a pressing need \cite{alzahrani2024benchmarkstargetsrevealingsensitivity,gan2024reasoningrobustnessllmsadversarial,li2024opensourcedatacontamination} for \emph{direct evaluation of problem-solving} capabilities as they are better aligned to the actual use-cases of LLMs. For example, while the GPQA dataset~\cite{rein2023gpqagraduatelevelgoogleproofqa} contains multiple-choice questions~(MCQs) that indirectly test the graduate-level knowledge of the models, the corresponding real-world use-case is more subjective -- the model should possess the ability to have an accurate open-ended conversation with the user about advanced concepts. This ability is better reflected in the strategies that the LLM uses to answer GPQA questions. Hence, the focus of this paper is to directly evaluate the internal problem-solving capabilities of the models in a holistic, yet scalable and generalizable manner.

We propose an evaluation framework called \emph{cascaded information disclosure} which focuses on eliciting and estimating the problem-solving capabilities reflected in the intermediate traces of the models. It involves modification of standard QA tasks as follows. We first transform the question into its more generalized form and then design a scheme to partition this form into \emph{non-overlapping parts}. Then we disclose these parts to the model under evaluation in a stagewise manner. While this framework accommodates multiple stages/parts, the final stage in all instantiations is a \emph{verifiable projection} stage that projects the model's responses in the earlier stages to an automatically evaluatable/verifiable form. This progressive question disclosure leads to a more accurate assessment of the subjective problem-solving capabilities of the models while the verifiable projection stage maintains the scalability and automation of evaluation. We instantiate this framework for two distinct QA modes: i) multiple-choice QA, and ii) reasoning-based math word problems.

Comparing a wide variety of models of different sizes and families~(\S~\ref{sec:experimental_setup:baselines_and_methods}), our approach narrows the performance gap observed in the standard evaluation settings, indicating that the prevalent indirect evaluation paradigm overestimates the differences in performance between models. Overall, we find that our approach not only provides a better estimate of LLMs' underlying reasoning capabilities, but also elicits better performance and intermediate traces from the models compared to the standard QA paradigm -- an effect primarily driven by the decomposition of the questions into separate foci. We further validate the soundness of our observations by extensive ablation studies.  

To summarize, we i) propose a novel generalized framework to test the problem-solving capabilities of LLMs that supports automation and extensibility, ii) empirically study our framework via concrete automatic instantiations across multiple QA tasks and LLMs, and iii) create manual instantiations of this framework by annotating two datasets that will be publicly available.

\section{Cascaded Information Disclosure} 
\label{sec:cid}
The central guiding conjecture \cite{wood1976role,collins1983cognitive} behind our approach is that an answerer (human or machine) will employ more detailed reasoning and knowledge-based strategies and reveal more information in their answers if the question is broken down into a series of leading questions. Therefore, our approach involves exposing partial information about the question to the answerer model in a cascade of stages and eliciting potentially detailed and meaningful responses from it at each stage. More concretely, the input question $Q$ is first converted into an abstract form $\bar{Q}$. The abstract question is then decomposed into $n$ non-overlapping subsets $\bar{Q} = \bigcup_{i=1}^n \bar{Q}_i$. Our approach then solicits responses $R_i$ from the model by presenting these subsets in a predetermined order and conditioning each response on the corresponding subset and the response in the previous stage: $R_i \leftarrow Q_i, R_{i-1}$. The response of the final stage is considered to be the model's answer to the full question: $A(Q) = R_n \mid Q_n, R_{n-1}$. Crucially, the response at any stage doesn't explicitly depend on the subparts of the question at the earlier stages -- the only information the model can gather about the previous stages is through the response generated one stage prior. This promotes detailed and informative responses from the model at each stage which, as we show in our results, is the primary driver for eliciting better problem-solving strategies from the models. An appropriate decomposition procedure underpins the success of our approach and needs to be custom-designed for every instantiation of this abstract framework.

In our experiments, we instantiate this framework for two different types of questions: multiple-choice questions, and math word problems. While the instantiations are procedurally different, there are some commonalities between the two instantiations. Both of them decompose the question into two stages (n=2) that we call \textbf{generalized ideation} and \textbf{verifiable projection} stages. The \emph{generalized ideation} stage presents a partial generalized view of the original question to the model such that the model generates a broad and general response $R_1$. The \emph{verifiable projection} stage takes this response along with the rest of the information in the original question (question residue) to project $R_1$ to an objectively measurable answer format. Therefore, the generalized ideation stage produces the subjective response detailing the problem-solving strategies employed by the models, and the verifiable projection stage ensures that the evaluation remains automatic and scalable by using standard evaluation metrics for the projected response. 

\subsection{MCQA Instantiation}
We break the MCQ into two parts: the question stem, and the options containing the answer. Therefore, the generalized ideation stage only gets to look at the general question without the knowledge of the answer candidate and induces the model to produce a detailed constructed response \cite{Livingston2009ConstructedResponseTQ} to answer the question in a free-form subjective manner. This response is then passed, along with \emph{only} the candidate options, to the verifiable projection stage. This stage assesses the model's free-form response and picks the option that most appropriately matches the response. 
We conduct experiments on different MCQA datasets designed to test for reasoning and advanced graduate-level knowledge.

\subsection{Math Word Problem Instantiation}
The original math word problems are rewritten into a templated form that is more general and specific entities are replaced by abstract variables. This generalized question is presented to the model in ideation stage in which the only correct way to answer the question is to reason about the problem and unknown variables, and produce a general mathematical formula for the generalized question. The verifiable projection stage takes the ideation response and the variable assignment as input. The mathematical formula is turned into executable code that accepts the variable assignments as arguments to yield a numeric answer to the original question. This instantiation requires considerable manual intervention. Therefore, we use the GSM8K~\cite{cobbe2021trainingverifierssolvemath} dataset for original questions whose generalized form is constructed using templates in prior work \cite{mirzadeh2024gsmsymbolicunderstandinglimitationsmathematical}. 

\section{Experimental Setup}



We empirically compare the effectiveness of the cascaded information disclosure evaluation framework against other prevalent frameworks by evaluating a \textbf{variety of open-weight models} spanning various sizes (7B to 32B) and model families (Llama \cite{grattafiori2024llama3herdmodels}, Qwen \cite{qwen2025qwen25technicalreport}, Gemma \cite{gemmateam2024gemma2improvingopen}, and Phi \cite{abdin2024phi4technicalreport}). Specifically, we use the instruction- finetuned  checkpoints for the models that are capable of following instruction prompts in a zero-shot QA setting. (Details in Appendix)

\noindent \textbf{Datasets:} As described in\S~\ref{sec:cid}, we instantiate our cascaded information disclosure framework for problem-solving evaluation for two types of questions: MCQs and math word problems. For the MCQA format, we picked a) the test split of ARC-Challenge \cite{clark2018thinksolvedquestionanswering} dataset which consists of grade school level science questions for MCQA format and, b) the GPQA dataset which consists of challenging graduate-level science questions written by domain experts. 
Specifically, we use the high-quality subset of the full GPQA dataset: GPQA-Main (448 questions), and its more challenging subset GPQA-Diamond (198 questions).
It is reported in the GPQA paper \cite{rein2023gpqagraduatelevelgoogleproofqa} that highly skilled non-expert human validators only reach 30.4\% and 22.1\% accuracies on GPQA-Main and GPQA-Diamond despite spending on average over 30 minutes with unrestricted access to the internet. 
For math word problems, we use GSM8K to obtain original questions. Previous work \citet{mirzadeh2024gsmsymbolicunderstandinglimitationsmathematical} created 100 symbolic templates from 100 questions in the original GSM8K dataset by manually annotating the entities (numeric values and categorical values) that can be abstracted into variables. We extend these templates by creating consistent assignments to the variables in the templates, resulting in a derivative dataset we call \emph{GSM-General}\footnote{
We will make this converted symbolic version of GSM8K dataset, \textbf{GSM-General} publicly available.} consisting of 100 template-variable assignment pairs that map to the 100 original questions in GSM8K dataset. 
 


\noindent \textbf{Inference Details}: For prompt engineering, we adopt zero-shot, chain-of-thought prompting \cite{wei2023chainofthoughtpromptingelicitsreasoning} to mimic the most common real-world LLM use case. Prompt templates are available in appendix \ref{appendix:prompt_templates}. 
We specifically crafted the prompts for all the settings including the baseline methods so as to achieve maximal performance on the tasks under each method. Specifically, we found that describing our cascaded setup for question disclosure to the models via the prompt resulted in stable behavior under our approach. We use XML tags based answer format in both prompting and answer extraction. 
We use greedy decoding which is the default setting for all three selected datasets. For each model, we set the context length to the maximum supported by the model and we set a uniform cap of 8192 for maximum number of generated tokens to facilitate a fair comparison across models.


\subsection{Evaluation Strategies Compared}
\label{sec:experimental_setup:baselines_and_methods}





\subsubsection{Standard Evaluation}
This evaluation setting is the most widely used evaluation method for the datasets we work with. We further enhanced the standard evaluation by developing custom prompts and answer extraction methods to ensure maximal performance on the datasets that would serve as difficult baselines for our approach to beat.
For \textbf{MCQA} type, the models take in the full original question: stem (\texttt{What is Newton's Third Law of Motion about?}) and candidate options (\texttt{A: Inertia, B: Force and acceleration, C: Momentum, D: Action-Reaction}) as inputs, and provide a response from which the answer index (\texttt{A, B, C, or D}) is extracted.\footnote{We use two implementations of standard MCQA baselines: LM Evaluation Harness  (\cite{eval-harness}; denoted ``LMH'') and a custom implementation. We found that LMH's simple prompts and answer extractions led to high parsing failure rate. 
To provide a fairer comparison of model performance, we develop an advanced MCQA baseline that greatly improves model's instruction following on formatting.
}

A \textbf{math word problem} is a mathematical exercise presented as a scenario, requiring the application of mathematical reasoning and computation to solve a real-world or hypothetical situation and provide a numerical answer. For evaluation, the prompt contains requirements on the format of the final answer. We use lm-evaluation-harness's implementation to evaluate models' performance on the 100 original GSM8K questions. 

\subsubsection{Cascaded Information Disclosure}
As described above, we instantiate both the question types with two stages: generalized ideation and verifiable projection. In the \textbf{generalized ideation stage}, we break the original question (or its transformed abstract form) into a \textit{generalized question} and \textit{question residue}.
In the MCQA setting, we construct the generalized question by concealing the options, which will then be used as question residue in the subsequent stage. As discussed in \S~\ref{sec:annotate}, we also experiment with obtaining the generalized question by concealing the options and manually rewriting the question stem to make it option-independent and general. In math word problem setting, we construct the generalized question by converting the original question into its symbolic form 
that replaces entities and quantities in the original question with abstract variables, leaving the value assignment to the variables involved as the question residue. In the prompt for generalized ideation stage, we explicitly specify that reiteration of the question is prohibited to restrict its leakage into the next stage. 

In the \textbf{verifiable projection stage}, we project the generalized reasoning traces in the ideation stage to either an option in MCQA or a numeric result in math word problem through a projector. The projector for MCQA tasks is an LLM whose task is to pick the most relevant option based solely on evaluation subject's ideation stage traces (no access to generalized question). We also explicitly instruct the projector LLM to abstain from solving the original problem and only focus on matching to the correct option to ensure that the accuracy on answers predicted by the projector provides a reliable estimate of the quality of the open-ended reasoning trace generated in the ideation stage. Similarly, the projector for the math word problem is only provided the trace from the ideation stage and is tasked with producing an executable Pythonic expression of the logic and formula expressed in the ideation stage. This code is then executed with the variable values to verify the correctness of the answer produced. We experiment with different kinds of projectors: a) the LLM being evaluated as the projector (self), b) an external performant open-weight projector (model Phi), c) a blackbox frontier projector model GPT-4o \cite{openai2024gpt4ocard}, and d) a rule-based projector that doesn't use any LLMs.
For rule-based substitute of LLM projector in MCQA, we implement a sentence-level BLEU score metric \cite{post-2018-call} for matching the first stage traces to the ground-truth answer. In the math word problem setting, we implement a function that extracts the expression stated in \emph{answer} tags from the ideation stage output. If executable, this expression is used to produce the answer by filling variable values from the question residue.

\subsubsection{Subjective evaluation of Generalized Ideation Traces} While verifiable projection stage makes our approach automatic and reliable, it is not the only possible automatic approach to evaluate the problem-solving capabilities of the model elicited in the generalized ideation stage. We also compare our setup against strategies for automatic subjective evaluation of the first stage traces. We experiment with \textbf{LLM-as-Judge projector}~\cite{zheng2023judgingllmasajudgemtbenchchatbot} for both MCQA and math word problem settings. The LLM judge, armed with the ground truth answer, is tasked to project the ideation trace to a binary verdict "Correct" and "Incorrect". 
 

\subsection{Evaluation Metrics}
Following the description above, we use the term \emph{accuracy} to refer to two kinds of metrics whose usage is differentiated by context: a) objective accuracy, which is a measure of match between the ground truth answer and the projected answer, and b) subjective accuracy, which is the rate of correctness as judged by an LLM. Thus, the standard baselines and systems employing the verifiable projector use objective scoring, while LLM-as-a-judge methods use subjective scoring
that decides whether the ideation trace is deemed correct or not. 
Subjective accuracy is heavily biased towards judge's style and hence is not directly comparable to objective accuracy.
We also report the parsing failure rate (in Appendix) of answer extraction. The failure is caused by the answering model's inability to follow answer formatting instructions.

\section{Empirical Analysis}
\subsection{Cascaded Information Disclosure Narrows the Gap between Models} 

\begin{table*}[htbp]
    \centering
    \caption{
    Comparison of Cascaded Information Disclosure based evaluation with verifiable projector with standard (Std.) evaluation. Experiments with three different kinds of projectors are reported: the LM under evaluation (self), external open-weight LM (Phi-4), and blackbox frontier model (GPT-4o). Performance Gap refers to the difference between the best performing model and the worst performing model.
    }
    \label{tab:experimental_results}
    \setlength{\tabcolsep}{2.5pt}
\begin{tabular}{|c|ccc|cccc|cccc|ccc|}
\hline
Dataset $\rightarrow$            & \multicolumn{3}{c|}{ARC-challenge}                     & \multicolumn{4}{c|}{GPQA-Diamond}                                        & \multicolumn{4}{c|}{GPQA-Main}                                           & \multicolumn{3}{c|}{GSM8K-100}                         \\ \hline
Model $\downarrow$             & \multicolumn{1}{l|}{} & \multicolumn{2}{c|}{Projector} & \multicolumn{1}{l|}{} & \multicolumn{3}{c|}{Projector}                   & \multicolumn{1}{l|}{} & \multicolumn{3}{c|}{Projector}                   & \multicolumn{1}{l|}{} & \multicolumn{2}{c|}{Projector} \\ \hline
                   & Std.                  & Self           & Phi-4         & Std.                  & Self           & Phi-4          & GPT-4o         & Std.                  & Self           & Phi-4          & GPT-4o         & Std.                  & Self           & Phi-4         \\
Llama 3.1-8B       & 82.5                & 80.4          & 87.7         & 20.7                & 31.8          & 32.8          & 35.3          & 23.7                 & 28.3          & 33.0          & 30.8          & 32                    & 40             & 42            \\
Gemma 2-9B         & 90.4                 & 84.6          & 90.1         & 30.30                 & 31.3          & 31.8          & 35.3          & 32.4                 & 30.6          & 32.6          & 34.8          & 21                    & 70             & 74            \\
Qwen 2.5-7B        & 87.2                 & 82.8          & 89.0         & 32.32                 & 30.3          & 34.8          & 37.4          & 31.0                 & 31.9          & 30.1          & 29.2          & 76                    & 75             & 70            \\
Qwen 2.5-14B       & 92.1                 & 88.8          & 90.9         & 41.4                 & 40.4          & 41.9          & 37.9          & 38.2                 & 36.2          & 37.7          & 37.3          & 80                    & 85             & 84            \\
Phi-4 (14B)        & 95.7                 & 90.8          & 90.8         & 58.1                  & 51.5         & 51.5          & 47.5          & 52.9                  & 45.3          & 45.3          & 47.1          & 52                    & 91             & 91            \\
Qwen 2.5-32B       & 93.3                 & 91.0          & 92.9         & 47.0                 & 42.4          & 44.9          & 46.5          & 43.1                 & 41.5          & 40.8          & 42.6          & 87                    & 95             & 95            \\
\textbf{Perf. Gap $\Delta$} & \textbf{13.2}        & \textbf{10.7} & \textbf{5.2} & \textbf{37.4}        & \textbf{21.2} & \textbf{19.7} & \textbf{12.1} & \textbf{29.2}        & \textbf{16.9} & \textbf{15.2} & \textbf{17.9} & \textbf{66}           & \textbf{55}    & \textbf{53}   \\ \hline
\end{tabular}
   
\end{table*}

In Table \ref{tab:experimental_results}, we present accuracies of LLMs evaluated on four different benchmark datasets using various evaluation paradigms. We report three different LLM projectors for the verifiable projection stage: Phi-4, GPT-4o, and ideation model itself. For each evaluation method, we measure performance gap as the difference between the highest and the lowest scoring model.
We observe that our approach results in a significantly narrower gap between the models compared to the standard evaluation. For example, the gap under standard evaluation for GPQA-Diamond~($37.4$) is much higher than the gap~($12.1$) when GPT-4o is used as the verifying projector. Moreover, the more powerful the verifying projector, the lower the performance gap.   
This suggests that the standard evaluation scheme prevalent today overestimates the difference between language model's capabilities.

\subsection{Cascaded Information Disclosure Estimates Problem-solving Capability}
\noindent \textbf{Generalized questions make QA tasks harder:}
\label{sec:harder_qa}
The generalized question used in our framework is more difficult than the original questions, by their nature. 
For MCQA, the answer choices are omitted in the generalized questions, thereby providing the evaluation subjects with strictly less information compared to standard evaluation methods. 
For GSM-General, the question is transformed into an abstract symbolic question, forcing the models to utilize their reasoning capabilities. 

The increase in difficulty is naturally reflected by decrease in performance of larger models (14B+ parameters) on MCQA tasks in Table~\ref{tab:experimental_results} under our evaluation paradigm. A qualitative example demonstrating this is provided as Phi-4's raw reasoning traces under standard MCQA ~\ref{lst:arc:standard} and cascaded information disclosure ~\ref{lst:arc:cid} for question \#996 in ARC-Challenge dataset. Under standard MCQA setting where options are provided, Phi-4 only analyzes the four options to get the correct answer whereas in our evaluation setting, Phi-4 falsely identified "nucleus" as not being part of the atom during ideation and its ideation trace is projected to the \textbf{Incorrect} answer by our verifiable projector correctly.
The increase in difficulty posed by our framework forces models to engage with the task at a deeper level, thereby contributing to superior evaluation of the model's problem-solving capabilities. 

\noindent \textbf{Generalized questions elicit better traces from the models:}
\label{sec:better_reasoning_trace}
By contrast, we observe the opposite trend for the smaller models (<10B) and the GSM8K dataset: the model performance is significantly better under our framework with verifiable projection than the standard evaluation setting. 
In fact, stronger projectors yields better performance.
This is caused by the improvement in the reasoning traces generated by the models under our evaluation framework, as we observe from manual inspection. 
In one of the GSM8K outputs (Listing~\ref{lst:gsm8k}), we observe that the Gemma-9B's outputs are incoherent past the third sentence for the standard evaluation method. In contrast, the reasoning traces for the projector method are semantically coherent, laying out each step coherently and explicitly. 
We observe similar behavior for GPQA outputs. 
In the standard evaluation method (Listing~\ref{lst:custom:qwen}), Qwen-7B's reasoning traces for a quantum mechanics problem are correct up to the middle, after which it hallucinates its expression to match one of the provided answer choices. For the projection method (Listing~\ref{lst:ideation:qwen}), the model generates a correct reasoning trace from start to finish, without any hallucinations. 

Our framework elicits higher quality traces from the model, thereby providing a more faithful estimation of its problem-solving capability.

\subsection{Separation of Concerns Improves Instruction Following Quality}
\label{sec:parsing_failure}



Recent work on RLVF has shown that improvements in MCQA are often driven not by enhanced reasoning, but by better output formatting capabilities introduced via fine-tuning or reinforcement learning \cite{shao2025spuriousrewardsrethinkingtraining}. This suggests that standard evaluations underestimate models' problem solving abilities due to brittle evaluation pipelines that rely on rigid regex-based answer extraction.

\noindent To quantify the effect of formatting issues, we compare parsing failure rates (full table in Appendix Table~\ref{tab:formatting:parsing_failure_rates}) between our two-stage method with Phi-4 as the final projector and the standard evaluation paradigm. For standard evaluation paradigm, we compare three systems: LMH (widely used LM evaluation harness), LMH+ (LM evaluation harness augmented with additional formatting and parsing instructions), and custom prompting that we designed specifically for the standard MCQA setup \ref{sec:experimental_setup:baselines_and_methods}. We observed that LMH and LMH+ evaluation setups are completely unsuitable for the Phi-4 model (parsing failure rate of 90\%+ on GPQA). This is because lm-evaluation-harness evaluation system doesn't apply any chat template even for instruction-finetuned models and Phi-4 tends to produce empty output when the input is not aligned with its prompt template. 
In general, we observe that the standard widely used LMH system suffers from high answer parsing error rates for a majority of our models: upwards of 40\% and 30\% for Gemma2-9B and the Llama3.1-8B models, and around 2-5\% for the Qwen models. LMH+ reduces these failure rates but they remain significantly high for the Llama and Gemma models (15-20\%). This indicates that models struggle to adhere to the implicit formatting assumptions of the lm-evaluation-harness framework.  
Our custom standard method which applies correct chat template for each model via careful prompt engineering and answer extraction strategies detailed in \ref{sec:experimental_setup:baselines_and_methods} further reduces parsing failures to below 10\% for most models except the Llama model, possibly because of length issues.

Contrastingly, our evaluation setting achieves \textbf{zero parsing failures} across all models. This is expected -- decomposition of the answering mechanism into ideation and projection stages enables disentangled focus on each stage leading to greater format adherence. 
More importantly, the reduction (or elimination) of formatting issues with custom-standard and our approach indicates that the performance trends in Table~\ref{tab:experimental_results} are an accurate reflection the models' performance on QA tasks, having accounted for a commonly observed 
confounding factor related to formatting.
\subsection{Oracle for Generalized Ideation}
\label{sec:oracle}

We evaluate the quality of two projection strategies in our framework: a) LLM-judge projector, b) verifiable projector. Specifically for GPQA, we bypass the LLM-based ideation stage and instead use human expert-written explanations as inputs to the projection stage. These explanations are provided by the expert annotators and are included as part of the GPQA dataset.
Thus, in this setting the LLM-judge projector should score 100\% because all the explanations are correct by design. Also, a high-quality verifiable-projector should always pick the correct option congruent with the ground-truth explanation.
\begin{table}[htbp]
\setlength{\tabcolsep}{3.5pt}
\caption{Results of subjective projection using LLM-as-judge (Judge) and verifiable projection (Verify) of oracle generalized ideation on GPQA-Main. Different projector LLMs are compared: Llama~3.1-8B(L), Gemma~2-9B (G), Qwen 2.5 models (Q-7B, Q-14B, Q-32B), and Phi-4 (P).
    }
    \label{tab:oracle_ideation}
\begin{tabular}{|ccccccc|}
\hline
       & L  & G  & Q-7B & Q-14B & P & Q-32B \\ \hline
Judge  & 65.6        & 95.3      & 70.1       & 92.4        & 97.3        & 93.1        \\
Verify & 95.8        & 95.5      & 97.8       & 99.1        & 98.2        & 99.3       \\ \hline
\end{tabular}
\end{table}
\noindent The judge setting in \noindent Table~\ref{tab:oracle_ideation} assumes access to additional knowledge of the correct answer to the original question to subjectively evaluate the oracle. We see that the judge performs poorly for several LMs (It is even poorer when it doesn't have access to original question's answer -- full table: Appendix table~\ref{tab:quality_of_projector}). This indicates that subjective evaluation techniques like LLM-as-a-judge do not faithfully reflect the assessment of problem-solving capabilities of models. Similar concerns about this evaluation paradigm have been raised in prior work~\cite{panickssery2024llmevaluatorsrecognizefavor,zheng2023judgingllmasajudgemtbenchchatbot,dubois2025lengthcontrolledalpacaevalsimpleway,zhu2025judgelmfinetunedlargelanguage,li2023haluevallargescalehallucinationevaluation,manakul2023selfcheckgptzeroresourceblackboxhallucination,doddapaneni2024findingblindspotsevaluator}.
In contrast, the verifiable projector achieves near-perfect accuracy across \emph{all} models, reaching over 99\% in several cases. This validates the correctness of our verifiable projection approach as it behaves almost flawlessly with oracle generalized ideation.
\subsection{Scalability of Our MCQA Instantiation}
\label{sec:annotate}


One potential concern in our implementation of MCQA instantiation of cascaded information disclosure framework is that some multiple-choice questions may not be fully self-contained once the candidate options are removed. Therefore, for challenging GPQA-Diamond questions, we projected the original MCQ~($Q$) to a modified abstract form~($\bar{Q}$) by \emph{manually annotating} the question stems such that they are self-contained. Table~\ref{tab:annotate} shows the difference between performance under our evaluation approach with Phi-4 as the projector on the annotated dataset and the unmodified dataset. We only notice small differences in either direction across LLMs. 
These results indicate that \textbf{manual annotation is not necessary} for effective cascaded information disclosure evaluation. 
This finding reinforces the practical scalability of our approach, allowing researchers to repurpose existing MCQA datasets for improved evaluation without incurring significant human annotation costs.
\begin{table}[htbp]
\setlength{\tabcolsep}{3.5pt}
\caption{Performance difference ($\Delta$) between two strategies for generalized question generation for MCQA: a) converting GPQA-diamond question stems to be self-contained, and b) using the original question stems. These questions are input to the ideation stage with different LLMs: Llama~3.1-8B(L), Gemma~2-9B(G), Qwen 2.5 models(Q-7B, Q-14B, Q-32B), and Phi-4(P).
    }
    \label{tab:annotate}
\begin{tabular}{|ccccccc|}
\hline
       & L  & G  & Q-7B & Q-14B & P & Q-32B \\ \hline
Annot.-$\Delta$  & 0.0        & 3.5      & -2.0       & -5.6        & 3.5        & -1.0        \\ \hline
\end{tabular}
\end{table}

\subsection{Effect of varying projector LLM}

\begin{table}[htbp]
\setlength{\tabcolsep}{2.4pt}
\caption{Performance Gap between the highest scoring projector LLM and the lowest scoring projector LLM for both subjective projector ($\Delta_{\textrm{Judge}}$) and verifiable projector ($\Delta_{\textrm{Verify}}$) settings on ARC-challenge~(A), GPQA-main~(M) and diamond~(D) datasets. The ideation stage is reported under different LLMs: Llama~3.1-8B(L), Gemma~2-9B (G), Qwen 2.5 models (Q-7B, Q-14B, Q-32B), and Phi-4 (P).
    }
    \label{tab:projLLM}
\begin{tabular}{|ccccccc|}
\hline
       & L  & G  & Q-7B & Q-14B & P & Q-32B \\ \hline
$\Delta_{\textrm{Judge}}$~(M)  & 9.6        & 27.0      & 12.7       & 19.9        & 16.7       & 18.3        \\ 
$\Delta_{\textrm{Verify}}$~(M)  & 4.7        & 3.1      & 2.2       & 3.6        & 5.8        & 4.2        \\\hline
$\Delta_{\textrm{Judge}}$(D)  & 11.6        & 23.7      & 12.1       & 19.19        & 15.1        & 17.7        \\ 
$\Delta_{\textrm{Verify}}$(D)  & 2.0        & 6.6      & 8.1       & 10.1        &  7.1        & 7.6        \\\hline
$\Delta_{\textrm{Judge}}$(A)  & 18.3        & 31.9      & 15.5       & 11.5        & 8.2        & 9.8        \\ 
$\Delta_{\textrm{Verify}}$(A)  & 7.3        & 7.6      & 6.9       & 6.7        & 6.9        & 6.6        \\\hline
\end{tabular}
\end{table}
We have shown in \ref{sec:oracle} that verifiable projectors consistently outperform subjective projectors when ground truth explanations are used as ideation traces. To further compare the robustness of subjective and verifiable projectors, we compute the accuracy gap between the best projector and the worst projector when different LLM projectors are used to score the ideation traces from the same models. From the results table \ref{tab:projLLM} we can see that this gap is consistently larger for subjective projectors than verifiable projectors. In some cases, LLM-as-a-Judge based projector performance gap is almost 7 times of verifiable projector, signaling extreme instability of the subjective projectors.
\subsection{Why LLMs for Verifiable Projection?}



\begin{table}[htbp]
\setlength{\tabcolsep}{3pt}
\caption{Performance difference ($\Delta$) between two verifiable projection strategies: a) using an LLM and, b) using a rigid rule-based system. The ideation stage is produced by different LLMs: Llama~3.1-8B(L), Gemma~2-9B (G), Qwen 2.5 models (Q-7B, Q-14B, Q-32B), and Phi-4 (P), on MCQA (GPQA) diamond and math word problem (GSM)8K-100 datasets.
    }
    \label{tab:rule}
\begin{tabular}{|ccccccc|}
\hline
       & L  & G  & Q-7B & Q-14B & P & Q-32B \\ \hline
GPQA~$\Delta$  & 12.1        & -1.0      & 6.1       & 6.6        & 17.2        & 11.62        \\ 
GSM~$\Delta$  & 12        & 15      & 22       & 9        & 0        & 0        \\\hline
\end{tabular}
\end{table}
\noindent Since the verifier projection stage maps the ideation stage output to an objective answer, it is possible to build simple heuristics for this stage instead of using an LLM as a projector. For MCQA, the rule-based projector selects the option that maximizes sentenceBLEU with the ideated response. For GSM8K, we extract a pythonic expression if available from the tail of the ideated response and execute it for evaluation.
Table~\ref{tab:rule} compares LLM-based projector~(Phi-4) to rule-based projectors designed by us (full table in Appendix Table \ref{tab:rule_based_accuracy}).  
Across almost all models and datasets, the LLM projector substantially outperforms ($+$~$\Delta$) the symbolic baseline. 
Thus, rigid rule-based systems are insufficient for reliable projection, particularly in tasks requiring nuanced reasoning alignment. On the other hand, this projection task is simple enough for LLM projectors as shown in \S~\ref{sec:oracle}. 

\subsection{Is It Just Inference-time Scaling?}
Because, our framework involves more computation steps due to its stagewise nature, in Table~\ref{tab:compute} we compare it (self as a verifier projector) to a two-round version of the standard evaluation setting~(self-reflect). This setting performs the standard QA twice, with the second round also using the model's response from the first round. We find that self-reflect behaves in an opposite manner to our framework -- it hurts the performance on the small models and doesn't affect the performance of larger models, thereby \emph{increasing} the performance gap between the models. This suggests that cascaded information disclosure performance isn't merely an artifact of inference-time scaling. 
\begin{table}[htbp]
\setlength{\tabcolsep}{2.5pt}
\caption{Performance gap between Increased-compute and Standard settings. Two methods of increasing compute are compared: a) cascaded info disclosure with the ideation LLM as verifiable projector ($\Delta_{\textrm{Ours}}$), and b) self-reflection ($\Delta_{\textrm{reflect}}$) on GPQA-Main~(M) and Diamond~(D). The ideation stage is reported under different LLMs: Llama~3.1-8B(L), Gemma~2-9B (G), Qwen 2.5 models (Q-7B, Q-14B, Q-32B), and Phi-4 (P).
    }
    \label{tab:compute}
\begin{tabular}{|ccccccc|}
\hline
       & L  & G  & Q-7B & Q-14B & P & Q-32B \\ \hline
$\Delta_{\textrm{Ours}}$~(M)  & 4.7        & -1.8      & 0.9       & -2.0        & -7.6        & -1.6        \\ 
$\Delta_{\textrm{reflect}}$~(M)  & -4.9        & 0.4      & -0.7       & -0.5        & 1.6        & 0.0        \\\hline
$\Delta_{\textrm{Ours}}$~(D)  & 11.1        & 1.0      & -2.0       & -1.0        & -6.6        & -4.6        \\ 
$\Delta_{\textrm{reflect}}$~(D)  & -2.5        & -1.5      & -1.0       & -1.5        & 1.0        & -2.5        \\\hline
\end{tabular}
\end{table}

\section{Discussion}

Similar to our instantiation, prior work has explored transforming extant benchmarks for better evaluation. This transformation includes permutation of answer choices \cite{eval-harness}, augmenting an additional choice \cite{wang2024mmluprorobustchallengingmultitask}, removing questions from the problem statement \cite{balepur2024artifactsabductionllmsanswer}, adversarial perturbations of problem statements \cite{li2024gsmpluscomprehensivebenchmarkevaluating,wang2022adversarialgluemultitaskbenchmark},translating the questions into another natural language \cite{yue2023mammothbuildingmathgeneralist}, formal language \cite{tsoukalas2024putnambenchevaluatingneuraltheoremprovers}, and programming language \cite{zheng2024codegeexpretrainedmodelcode}, among others. 
Another popular transformation for MCQA is randomly replacing the answer choices with \texttt{None of the other choices} \cite{balepur2025bestdescribesmultiplechoice,Livingston2009ConstructedResponseTQ}
We also conducted experiments with this transformation (Appendix Table \ref{tab:ablation_swap_accuracy}) and observed that this further shrinks the gap between the models as it makes the questions more difficult. Such transformations are complementary to our findings and can be seamlessly integrated into our framework.  

Finally, our evaluation framework is general and capable of encompassing multiple stages with distinct foci. Our observation that our multi-stage setup elicits improved intermediate traces suggests a fruitful future direction for \emph{learning} the decomposition \cite{prasad2024adaptasneededdecompositionplanning, Renze_2024} and nature of stages for enhanced problem-solving behavior via Reinforcement Learning (RL)  \cite{deepseekai2025deepseekr1incentivizingreasoningcapability, wen2025reinforcementlearningverifiablerewards}.  

\section{Limitations}

Although our proposed method provides a more robust evaluation of a model's reasoning capabilities, it requires some additional compute to run the LLM-based projector. Relatedly, our framework could be extended with recent orthogonal approaches for inference-time scaling \cite{wang2023selfconsistencyimproveschainthought}, \cite{madaan2023selfrefineiterativerefinementselffeedback}, \cite{snell2024scalingllmtesttimecompute, tian2025thinktwiceenhancingllm}. In this work, we focus on a structured and efficient multi-round paradigm that avoids redundant computation.

Furthermore, we restricted our attention to MCQA and math word problems in this paper. We did not evaluate on many other tasks of interest to the LLM community, such as machine translation. 
While our framework is general enough to accommodate multiple tasks and stages, for some of these cases, formulation and implementation may be nontrivial when instantiating our cascading information disclosure framework. For example, some of these cases might require annotating the dataset as in our GSM-General (although this process could be automated using an LLM).

\bibliography{custom}

\clearpage          
\section{Appendix}
\onecolumn

\subsection{Qualitative Examples}

\subsubsection{Example from GSM8K Dataset}





  \begin{lstlisting}[
  float,
  floatplacement=!htbp,
  % language=javascript,
  caption={GSM8K Example. Evaluation subject is Gemma-2-9B},
  label={lst:gsm8k}
]
"original_question": "Brett is 14 years old. In four years his sister Angela will be three times as old as he is now. How old is Angela right now?", 
"answer_numeric": 38, 
"generalized_symbolic_question": "{name1} is {age1} years old. In {years} years his {relation_type} {name2} will be {mult} times as old as {name1} is now. How old is {name2} right now?\n\nVariables and the range of possible values they can take are:\n- name1 = sample(names_male)\n- name2 = sample(names_female)\n- relation_type = sample(['sister', 'cousin'])\n- $age1 = range(8, 25)\n- $years = range(2, 10)\n- $mult = range(2, 5)\n\nThe relationship variables should have is:\n- age1 * mult - years > 0\n\n", 
"answer_symbolic" : "age1*mult-years", 
"ideation_traces": "<Reasoning>\n1.  Let's denote the current age of {name2} as {age2}.\n2.  In {years} years, {name1} will be {age1} + {years} years old.\n3.  In {years} years, {name2} will be {age2} + {years} years old.\n4.  We know that in {years} years, {name2} will be {mult} times as old as {name1} is now, so we can write the equation: {age2} + {years} = {mult} * {age1}\n5.  We need to solve this equation for {age2}.\n\n</Reasoning>\n<Answer>{age2} = {mult} * {age1} - {years}</Answer>",
"raw_symbolic_answer": "{age2} = {mult} * {age1} - {years}", 
"cleaned_symbolic_answer": "(mult * age1) - years",
"projected_numeric_result": 38, 
"lmh_answer_trace": "We need to figure out how old Angela is right now.  We know that in four years her sister Angela will be three times as old as he is now.  We also know that Angela is right now.  We need to figure out how old Angela is right now.\n\n\nLet me know if you have any other questions.<end_of_turn>", 
"lmh_filtered_answer": "[invalid]"
  \end{lstlisting}

\clearpage
\subsubsection{Example from ARC-Challenge Dataset}

  \begin{lstlisting}[
  float,
  floatplacement=!htbp,
  % language=javascript,
  caption={ARC-Challenge Example: Phi-4 Correctly Answers Question in Standard MCQA Setting},
  label={lst:arc:standard}
]

   "id": "996",
   "question": "Which of these is not part of an atom?",
   "options": [
     "isotope",
     "proton",
     "nucleus",
     "electron"
   ],
   "answer_index": 0,
   "Phi-4_raw_output": "<Reasoning>\nTo determine which option is not part of an atom, let's analyze each choice:\n\nA) Isotope: An isotope refers to variants of a particular chemical element that have the same number of protons but different numbers of neutrons. Isotopes are not a part of an atom itself; rather, they are different forms of atoms of the same element. Therefore, isotopes are not a component of a single atom.\n\nB) Proton: Protons are positively charged particles found in the nucleus of an atom. They are a fundamental part of the atom's structure.\n\nC) Nucleus: The nucleus is the central part of an atom, containing protons and neutrons. It is a crucial component of the atom.\n\nD) Electron: Electrons are negatively charged particles that orbit the nucleus of an atom. They are essential parts of the atom's structure.\n\nBased on this analysis, the option that is not a part of an atom is A) isotope, as it refers to different forms of atoms rather than a component of a single atom.\n</Reasoning>\n<PickedAnswer>A</PickedAnswer>",
   "is_correct": "Yes"
  \end{lstlisting}

  \begin{lstlisting}[
  float,
  floatplacement=!htbp,
  % language=javascript,
  caption={ARC-Challenge Example: Phi-4 Fails to Derive the Correct Answer in Cascaded Information Disclosure Setting},
  label={lst:arc:cid}
]

"id": "996",
   "phi-4_ideation_traces": "To determine which component is not part of an atom, we need to understand the basic structure of an atom. An atom is the smallest unit of ordinary matter that forms a chemical element. It consists of three primary subatomic particles:\n\n1. **Protons**: These are positively charged particles found in the nucleus of an atom. The number of protons in the nucleus defines the atomic number of an element and determines the element's identity.\n\n2. **Neutrons**: These are neutral particles, meaning they have no charge, and are also located in the nucleus alongside protons. Neutrons contribute to the mass of the atom and can influence the stability of the nucleus.\n\n3. **Electrons**: These are negatively charged particles that orbit the nucleus in various energy levels or electron shells. Electrons are much smaller in mass compared to protons and neutrons and are responsible for the chemical properties of an atom.\n\nGiven this understanding, any component not fitting into these categories (protons, neutrons, or electrons) would not be considered part of an atom. Commonly, other particles or structures that are not part of an atom include:\n\n- **Molecules**: These are groups of two or more atoms bonded together. Molecules are not part of a single atom but rather a collection of atoms.\n\n- **Ions**: These are atoms or molecules that have gained or lost electrons, resulting in a net charge. While ions are related to atoms, they are not a fundamental part of an atom's structure.\n\n- **Nucleus**: While the nucleus is a part of an atom, it is not a subatomic particle itself but rather a structure within the atom that contains protons and neutrons.\n\n- **Quarks**: These are fundamental particles that make up protons and neutrons. While quarks are part of the subatomic particles, they are not considered part of the atom itself in the context of basic atomic structure.\n\n- **Photons**: These are particles of light and do not form part of an atom. They are involved in electromagnetic interactions but are not components of atomic structure.\n\nBased on this analysis, the correct answer would be a component that does not fit into the categories of protons, neutrons, or electrons, such as a molecule, ion, nucleus, quark, or photon.",
"phi-4_ideation_stage_answer": "A molecule, ion, nucleus, quark, or photon is not part of an atom.",
"verifiable_projector_raw_output": "<Reason>\nThe key conclusions from the 'Reasoning Trace' are that an atom consists of three primary subatomic particles: protons, neutrons, and electrons. Any component that does not fit into these categories is not considered part of an atom. The trace specifically identifies molecules, ions, the nucleus, quarks, and photons as examples of components that are not part of an atom.\n\nNow, let's evaluate each answer option:\n\n- **Option A: proton**\n  - Protons are explicitly mentioned in the reasoning trace as a fundamental component of an atom. Therefore, this option does not match the conclusion that the component is not part of an atom.\n\n- **Option B: electron**\n  - Electrons are also explicitly mentioned in the reasoning trace as a fundamental component of an atom. Thus, this option does not match the conclusion that the component is not part of an atom.\n\n- **Option C: nucleus**\n  - The reasoning trace states that the nucleus is not a subatomic particle itself but a structure within the atom that contains protons and neutrons. Therefore, the nucleus is not considered a fundamental component of an atom in the context of subatomic particles. This aligns with the conclusion that the component is not part of an atom.\n\n- **Option D: isotope**\n  - Isotopes are variants of a particular chemical element that differ in neutron number. While isotopes are related to atoms, they are not a fundamental part of an atom's structure. However, the reasoning trace does not explicitly mention isotopes as an example of a component not part of an atom. The focus is more on subatomic particles and structures like the nucleus.\n\nBased on the analysis, Option C (nucleus) is the most consistent and closest match to the reasoning trace, as it is explicitly identified as not being a subatomic particle and thus not a fundamental component of an atom in the context of the trace.\n</Reason>\n<PickedAnswer>C</PickedAnswer>",
   "projected_answer_text": "nucleus",
   "is_anwer_correct": "No"
  \end{lstlisting}
\FloatBarrier

\subsubsection{Example from GPQA Dataset}
\textbf{Qwen2.5-7B's answer under standard MCQA Evaluation}




  \begin{lstlisting}[
  float,
  floatplacement=!htbp,
  % language=javascript,
  caption={Qwen Custom Example},
  label={lst:custom:qwen}
]

        "resps": "<Reasoning>\nTo solve this problem, we need to understand the Kraus representation of a quantum channel and how it relates to the given Kraus operators. The Kraus operators \\(A_i\\) for a depolarizing channel are given as:\n\n\\[ A_0 = \\sqrt{1 - \\frac{3p}{4}} I \\]\n\\[ A_1 = \\sqrt{\\frac{p}{4}} X \\]\n\\[ A_2 = \\sqrt{\\frac{p}{4}} Y \\]\n\\[ A_3 = \\sqrt{\\frac{p}{4}} Z \\]\n\nThe depolarizing channel can be represented as:\n\n\\[ E(\\rho) = \\sum_i A_i \\rho A_i^\\dagger \\]\n\nGiven the Kraus operators, we can substitute them into the equation:\n\n\\[ E(\\rho) = A_0 \\rho A_0^\\dagger + A_1 \\rho A_1^\\dagger + A_2 \\rho A_2^\\dagger + A_3 \\rho A_3^\\dagger \\]\n\nSince \\(A_0 = \\sqrt{1 - \\frac{3p}{4}} I\\), we have:\n\n\\[ A_0 \\rho A_0^\\dagger = (1 - \\frac{3p}{4}) \\rho \\]\n\nFor the other operators, we have:\n\n\\[ A_1 \\rho A_1^\\dagger = \\frac{p}{4} X \\rho X \\]\n\\[ A_2 \\rho A_2^\\dagger = \\frac{p}{4} Y \\rho Y \\]\n\\[ A_3 \\rho A_3^\\dagger = \\frac{p}{4} Z \\rho Z \\]\n\nCombining these, we get:\n\n\\[ E(\\rho) = (1 - \\frac{3p}{4}) \\rho + \\frac{p}{4} X \\rho X + \\frac{p}{4} Y \\rho Y + \\frac{p}{4} Z \\rho Z \\]\n\nWe can rewrite this as:\n\n\\[ E(\\rho) = (1 - \\frac{3p}{4}) \\rho + \\frac{p}{4} (X \\rho X + Y \\rho Y + Z \\rho Z) \\]\n\nWe know that \\(X^2 = Y^2 = Z^2 = I\\), so:\n\n\\[ X \\rho X = \\rho \\]\n\\[ Y \\rho Y = \\rho \\]\n\\[ Z \\rho Z = \\rho \\]\n\nThus, the expression simplifies to:\n\n\\[ E(\\rho) = (1 - \\frac{3p}{4}) \\rho + \\frac{p}{4} (\\rho + \\rho + \\rho) = (1 - \\frac{3p}{4}) \\rho + \\frac{3p}{4} \\rho = (1 - \\frac{3p}{4} + \\frac{3p}{4}) \\rho = (1 - p) \\rho + \\frac{p}{3} (X \\rho X + Y \\rho Y + Z \\rho Z) \\]\n\nThis matches the form given in option A.\n\nTherefore, the correct Kraus representation of the state \\(E(\\rho)\\) is:\n\n\\[ E(\\rho) = (1 - p) \\rho + \\frac{p}{3} (X \\rho X + Y \\rho Y + Z \\rho Z) \\]\n\n</Reasoning>\n<PickedAnswer>A</PickedAnswer>"
\end{lstlisting}




\FloatBarrier

\clearpage
\textbf{Qwen2.5-7B's ideation traces under our framework}:




  \begin{lstlisting}[
  float,
  floatplacement=!htbp,
  % language=javascript,
  caption={Qwen Ideation Traces},
  label={lst:ideation:qwen}
]
  
"reasoning": "To derive the Kraus representation of the depolarizing channel operation \( E(\rho) \), we need to ensure that the sum of the squares of the Kraus operators equals the identity matrix. The given Kraus operators are:\n\n\[ A_0 = \sqrt{1 - \frac{3p}{4}} I \]\n\[ A_1 = \sqrt{\frac{p}{4}} X \]\n\[ A_2 = \sqrt{\frac{p}{4}} Y \]\n\[ A_3 = \sqrt{\frac{p}{4}} Z \]\n\nFirst, let's verify that the sum of the squares of these operators equals the identity matrix \( I \):\n\n\[ A_0^2 + A_1^2 + A_2^2 + A_3^2 = \left( \sqrt{1 - \frac{3p}{4}} I \right)^2 + \left( \sqrt{\frac{p}{4}} X \right)^2 + \left( \sqrt{\frac{p}{4}} Y \right)^2 + \left( \sqrt{\frac{p}{4}} Z \right)^2 \]\n\nCalculating each term individually:\n\n\[ A_0^2 = \left( \sqrt{1 - \frac{3p}{4}} I \right)^2 = \left( 1 - \frac{3p}{4} \right) I \]\n\[ A_1^2 = \left( \sqrt{\frac{p}{4}} X \right)^2 = \frac{p}{4} X^2 = \frac{p}{4} I \]\n\[ A_2^2 = \left( \sqrt{\frac{p}{4}} Y \right)^2 = \frac{p}{4} Y^2 = \frac{p}{4} I \]\n\[ A_3^2 = \left( \sqrt{\frac{p}{4}} Z \right)^2 = \frac{p}{4} Z^2 = \frac{p}{4} I \]\n\nSumming these up:\n\n\[ A_0^2 + A_1^2 + A_2^2 + A_3^2 = \left( 1 - \frac{3p}{4} \right) I + \frac{p}{4} I + \frac{p}{4} I + \frac{p}{4} I \]\n\[ = \left( 1 - \frac{3p}{4} + \frac{p}{4} + \frac{p}{4} + \frac{p}{4} \right) I \]\n\[ = \left( 1 - \frac{3p}{4} + \frac{3p}{4} \right) I \]\n\[ = 1 I \]\n\[ = I \]\n\nSince the sum of the squares of the Kraus operators equals the identity matrix, the given Kraus operators correctly represent the depolarizing channel operation \( E(\rho) \).\n\nTherefore, the correct Kraus representation of the state \( E(\rho) \) is:\n\n\[ E(\rho) = \sum_{i=0}^{3} A_i \rho A_i^\dagger \]\n\nWhere the Kraus operators are:\n\n\[ A_0 = \sqrt{1 - \frac{3p}{4}} I \]\n\[ A_1 = \sqrt{\frac{p}{4}} X \]\n\[ A_2 = \sqrt{\frac{p}{4}} Y \]\n\[ A_3 = \sqrt{\frac{p}{4}} Z \]",
"answer": "\[ E(\rho) = A_0 \rho A_0^\dagger + A_1 \rho A_1^\dagger + A_2 \rho A_2^\dagger + A_3 \rho A_3^\dagger \]\nWhere \( A_0 = \sqrt{1 - \frac{3p}{4}} I \), \( A_1 = \sqrt{\frac{p}{4}} X \), \( A_2 = \sqrt{\frac{p}{4}} Y \), and \( A_3 = \sqrt{\frac{p}{4}} Z \)." 
  \end{lstlisting}

\FloatBarrier

\clearpage
\textbf{The projection traces in stage 2 - Phi-4 as projector}:




  \begin{lstlisting}[
  float,
  floatplacement=!htbp,
  % language=javascript,
  caption={projector=phi4},
  label={lst:projector:phi4}
]

"ideation_traces": "<Reason>\nThe reasoning trace provides a detailed derivation of the Kraus representation for a depolarizing channel. The key conclusions from the reasoning trace are:\n\n1. The Kraus operators for the depolarizing channel are:\n   - \( A_0 = \sqrt{1 - \frac{3p}{4}} I \)\n   - \( A_1 = \sqrt{\frac{p}{4}} X \)\n   - \( A_2 = \sqrt{\frac{p}{4}} Y \)\n   - \( A_3 = \sqrt{\frac{p}{4}} Z \)\n\n2. The sum of the squares of these Kraus operators equals the identity matrix, confirming they form a valid representation.\n\n3. The depolarizing channel operation \( E(\rho) \) is expressed as:\n   \[\n   E(\rho) = A_0 \rho A_0^\dagger + A_1 \rho A_1^\dagger + A_2 \rho A_2^\dagger + A_3 \rho A_3^\dagger\n   \]\n\n4. Substituting the Kraus operators into the expression for \( E(\rho) \), we get:\n   \[\n   E(\rho) = \left(1 - \frac{3p}{4}\right) \rho + \frac{p}{4} X \rho X + \frac{p}{4} Y \rho Y + \frac{p}{4} Z \rho Z\n   \]\n\nNow, let's evaluate each answer option:\n\n- **Option A:**\n  \[\n  E(\rho) = (1-p)\rho + \frac{p}{3}X\rho^{2}X + \frac{p}{3}Y\rho^{2}Y + \frac{p}{3}Z\rho^{2}Z\n  \]\n  This option uses \(\rho^2\) instead of \(\rho\) and has coefficients \(\frac{p}{3}\) instead of \(\frac{p}{4}\). It does not match the derived expression.\n\n- **Option B:**\n  \[\n  E(\rho) = (1-p)\rho + \frac{p}{4}X\rho X + \frac{p}{4}Y\rho Y + \frac{p}{4}Z\rho Z\n  \]\n  This option has the correct form and coefficients \(\frac{p}{4}\) for the terms involving \(X\), \(Y\), and \(Z\). However, the coefficient for the identity term is \((1-p)\) instead of \(\left(1 - \frac{3p}{4}\right)\). Despite this discrepancy, it is the closest match in terms of structure and coefficients.\n\n- **Option C:**\n  \[\n  E(\rho) = (1-p)\rho + \frac{p}{3}X\rho X + \frac{p}{3}Y\rho Y + \frac{p}{3}Z\rho Z\n  \]\n  This option uses \(\frac{p}{3}\) instead of \(\frac{p}{4}\) and \((1-p)\) instead of \(\left(1 - \frac{3p}{4}\right)\). It does not match the derived expression.\n\n- **Option D:**\n  \[\n  E(\rho) = (1-p)\rho^{2} + \frac{p}{3}X\rho^{2}X + \frac{p}{3}Y\rho^{2}Y + \frac{p}{3}Z\rho^{2}Z\n  \]\n  This option uses \(\rho^2\) instead of \(\rho\) and has incorrect coefficients \(\frac{p}{3}\). It does not match the derived expression.\n\nOverall, Option B is the closest match to the reasoning trace in terms of structure and coefficients, despite the slight discrepancy in the coefficient of the identity term.\n</Reason>\n<PickedAnswer>B</PickedAnswer>"

\end{lstlisting}

\FloatBarrier 

\subsection{Prompt Templates and XML Tag Based Answer Extraction}
\label{appendix:prompt_templates}

\subsubsection{Prompt Templates}
\textbf{Prompt used in our custom implementation of MCQA Evaluation}

  \begin{lstlisting}[
  float,
  floatplacement=!htbp,
  % language=javascript,
  caption={Custom Implementation of Standard MCQA Evaluation Prompt},
  label={lst:prompt:custom:mcqa}
]

You are a helpful assistant. You will be provided with a challenging science domain question (e.g., biology, physics, chemistry) and a set of multiple-choice options. Your task is to reason through the question step-by-step to arrive at a detailed answer and select the correct option.

Follow these steps:
1.  Think step-by-step to analyze the question and derive a solution. Detail your reasoning process.
2.  After your reasoning, clearly state your final answer by picking the correct option letter.

Output your response using the following XML format ONLY:
<Reasoning>
Your detailed step-by-step thinking process here.
</Reasoning>
<PickedAnswer>X</PickedAnswer> (where X is the letter of the correct option)

  \end{lstlisting}
\textbf{Prompt used for self-reflection}

  \begin{lstlisting}[
  float,
  floatplacement=!htbp,
  % language=javascript,
  caption={Self Reflection Prompt},
  label={lst:prompt:self_reflection}
]

You are an expert evaluator. You will be provided with a 'Reasoning Trace' from a language model that attempted to answer a question and multiple 'Answer Options' (labeled as A, B, C, etc.). The original question will also be provided.

Your task is to critically re-evaluate the entire problem to find the correct answer. You must perform your own independent, step-by-step analysis to arrive at the most accurate solution, using the provided 'Reasoning Trace' as a reference but not assuming it is correct.

Consider the following when making your decision:
-   **Numerical Closeness:** If the options and your reasoning involve numerical values, convert all relevant numbers to scientific notation (e.g., expressing numbers as a coefficient multiplied by 10 raised to an exponent, like 1.23 x 10^4 or 5.6 x 10^-2). Ensure all numbers are in this consistent format, and in the same unit if applicable, before comparison. Then, identify the option whose scientific notation value is closest to the scientific notation value derived from your reasoning.
-   **Conceptual Closeness:** If the options involve concepts, equations, definitions, or qualitative explanations, identify the option that best reflects the core ideas in your reasoning or is logically inferred from your reasoning.

Follow these steps:
1.  Carefully read and understand the original question, all options, and the entire 'Reasoning Trace'.
2.  Conduct your own detailed, step-by-step reasoning process to solve the problem. You can reference, correct, or build upon the provided trace.
3.  Based on your independent analysis, decide which option is the most consistent and closest match.

Output your evaluation using the following XML format ONLY:
<Reasoning>
Your detailed step-by-step reasoning here.
</Reasoning>
<PickedAnswer>X</PickedAnswer> (where X is the letter corresponding to your chosen option)

  \end{lstlisting}
\textbf{Prompt used in ideation stage}
\begin{lstlisting}[
  float,
  floatplacement=!htbp,
  % language=javascript,
  caption={Ideation Stage Prompt},
  label={lst:prompt:ideation}
]
You are a helpful assistant. You will receive a challenging science-domain question, but the multiple-choice options are hidden from you. Your step-by-step scientific reasoning will be passed to a Stage-2 AI that can see the options and will rely on your analysis to choose the correct one.

Therefore:
1. Do NOT refuse, disclaim, or say that the task is impossible because the options are missing.
2. Apply rigorous scientific reasoning: cite relevant concepts, principles, equations, variables, and alternative scenarios; deduce the most plausible answer or the exact characteristics the correct answer must have.
3. Do NOT quote, paraphrase, or repeat the question text in your response.

OUTPUT FORMAT --- STRICT
Respond using EXACTLY the XML structure shown below and nothing else.
Formatting rules:
- Put your full, detailed reasoning inside the <Reasoning> ... </Reasoning> tags.
- Put your concise final answer (or the uniquely identifying properties of the correct answer) inside the <Answer> ... </Answer> tags.
- Do not include any text before <Reasoning> or after </Answer>.
- Do not add extra tags, attributes, or wrapper elements.

<Reasoning>Your step-by-step thinking process goes here.</Reasoning>
<Answer>Your final derived answer goes here.</Answer>
\end{lstlisting}
\textbf{Prompt used by verifiable projector}
\begin{lstlisting}[
  float,
  floatplacement=!htbp,
  % language=javascript,
  caption={Verifiable Projector Prompt},
  label={lst:prompt:verifiable_projector}
]
You are an expert evaluator whose ONLY goal is to select the best answer OPTION (A, B, C, ...) based on a provided Reasoning Trace.

What you receive
- The Reasoning Trace: a step-by-step analysis already produced by another model.
- The Answer Options: labelled A, B, C, ... (the original question text may be omitted).

Your responsibilities
1. DO NOT solve the original question again. Judge options solely by how well they align with the Reasoning Trace's conclusions.
2. DO NOT refuse, claim uncertainty, or state that none of the options match. You MUST pick the single closest option even if the match is imperfect.
3. Use two dimensions of closeness:
   - Numerical closeness: convert every relevant number in both trace and options to scientific-notation in the SAME UNITS, then choose the option with the smallest numerical difference (consider both coefficient and exponent).
   - Conceptual closeness: match the key ideas, equations, or qualitative statements in the trace.
4. If several options tie, pick the first among the tied ones in alphabetical order and explain the tie-break.

OUTPUT FORMAT --- STRICT (XML)
You MUST comply with this exact structure; any deviation will cause automatic failure.
Rules
- Put your comparative analysis inside <Reason> ... </Reason>.
- End with a single uppercase letter (A, B, C, ...) inside <PickedAnswer> ... </PickedAnswer>.
- NEVER leave <PickedAnswer> blank, use words, or add extra tags/attributes.
- Do NOT output anything before <Reason> or after </PickedAnswer>.

<Reason>Your detailed comparison of each option to the Reasoning Trace goes here. Begin with a one-sentence summary of the trace's conclusion, then analyse each option (A, B, C...) explaining its fit or mismatch. Finish by justifying the option you will pick.</Reason>
<PickedAnswer>X</PickedAnswer>
\end{lstlisting}
\textbf{Prompt used by Subjective LLM Projector}

  \begin{lstlisting}[
  float,
  floatplacement=!htbp,
  % language=javascript,
  caption={LLM Judge Projector Prompt},
  label={lst:prompt:LLM_as_judge}
]

You are an expert evaluator. You will be provided with a 'Reasoning Trace' from a language model that attempted to answer a question. The original question may be provided, and in some cases, the correct answer may also be provided for reference.

Your task is to meticulously analyze the 'Reasoning Trace' and evaluate whether the reasoning process and conclusions are correct. You must not try to solve the original question yourself; your evaluation should be based solely on the logical soundness and accuracy of the reasoning presented.

Consider the following when making your decision:
-   **Logical Soundness:** Evaluate if the reasoning follows a clear, logical progression and if each step is properly justified.
-   **Numerical Accuracy:** If the reasoning involves numerical calculations, verify if the math is correct and if the units are properly handled.
-   **Conceptual Accuracy:** If the reasoning involves scientific concepts, equations, or definitions, verify if they are correctly applied and interpreted.
-   **Completeness:** Check if the reasoning addresses all aspects of the question and if any assumptions are clearly stated and justified.

Follow these steps:
1.  Carefully read and understand the entire 'Reasoning Trace'.
2.  If a reference answer is provided, use it to verify the accuracy of the reasoning.
3.  Identify any logical flaws, incorrect assumptions, or mathematical errors.
4.  Based on your analysis, determine if the reasoning is Correct or Incorrect.

Output your evaluation using the following XML format ONLY:
<Reason>
Your detailed step-by-step analysis here. First, summarize the key points in the reasoning trace. Then, explain any flaws or strengths you identified. Finally, state your overall justification for your judgment.
</Reason>
<Judgment>Correct</Judgment> or <Judgment>Incorrect</Judgment>

  \end{lstlisting}

\subsubsection{XML Tag Based Answer Extraction}
\label{appendix:XML tags}
As we shown above, we adopt a XML tag based prompting strategies where each functional components are wrapped with XML tag that summarizes its function. We also include clear instructions and a demonstration for the desired output XML format at the end. Then in the answer extraction stage, we use regular expression to search for the corresponding pattern (e.g. for answers, we are searching for $<$PickedAnswer$>$ and $<$/PickedAnswer$>$). In our experiments, we found that this method greatly improves the stability of models instruction following performance in answer format and results in a near zero parsing failure rates.



\subsection{Additional Experimental Results}

\begin{table*}[!htbp] 
    \centering
    \footnotesize
    
    \begin{subtable}{1\textwidth} 
        \centering
        \ra{1.3}
        \setlength{\tabcolsep}{3pt}
        \caption{Model accuracy on the GPQA Diamond dataset when the correct answer is swapped out with ``None of the other choices.'' For ``Phi (No swap)'', we report accuracies for the case when no changes are made to the answer choices. Higher is better. The $\Delta$ row reports the difference between minimum and maximum for each column. Parsing failure rates are 0\% for the projector methods.}
        \label{tab:ablation_swap_accuracy}
        \begin{tabular}{@{}lrrrr@{}}
            \toprule
            & & \multicolumn{3}{c}{Projector} \\
            \cmidrule(lr){3-5}
            & Standard & Self & Phi & Phi (No swap) \\
            \midrule
            \multicolumn{5}{@{}l}{\textbf{GPQA Diamond} - \textit{Accuracy (\%)}} \\
            Llama3.1-8B   & 15.66 & 28.28 & 37.37 & 32.83 \\
            Gemma2-9B     & 13.13 & 27.78 & 37.37 & 31.82 \\
            Qwen2.5-7B    & 27.78 & 31.82 & 41.41 & 34.85 \\
            Qwen2.5-14B   & 40.40 & 34.34 & 41.92 & 41.92 \\
            Phi-4 (14B)   & 3.03  & 45.96 & 45.96 & 51.52 \\
            Qwen2.5-32B   & 35.35 & 38.89 & 43.43 & 44.95 \\
            \midrule
            $\Delta$      & 37.37 & 18.18 &  8.59 & 19.7\\
            \bottomrule
        \end{tabular}
    \end{subtable}
    
    \vspace{1cm} 
    
    \begin{subtable}{1\textwidth} 
        \centering
        \ra{1.3}
        \setlength{\tabcolsep}{3pt}
        \caption{Parsing failure rates for the ablation studies on the GPQA Diamond dataset. The table compares failure rates when the correct answer is swapped with a "None" option versus when it is swapped with a random incorrect option.}
        \label{tab:ablation_swap_parsing_error}
        \begin{tabular}{@{}lrrrr@{}}
            \toprule
            & & \multicolumn{3}{c}{Projector (Ours)} \\
            \cmidrule(lr){3-5}
            Model & Standard & Self & Phi & Phi (No swap) \\
            \midrule
            \multicolumn{5}{@{}l}{\textbf{GPQA Diamond} - \textit{Parsing Failure (\%)}} \\
            Llama3.1-8B   & 29.29 & 0.00 & 0.00 & 0.00 \\
            Gemma2-9B     & 47.47 & 0.00 & 0.00 & 0.00 \\
            Qwen2.5-7B    &  5.56 & 0.00 & 0.00 & 0.00 \\
            Qwen2.5-14B   &  0.51 & 0.00 & 0.00 & 0.00 \\
            Phi-4 (14B)   & 91.41 & 0.00 & 0.00 & 0.00 \\
            Qwen2.5-32B   &  5.56 & 0.00 & 0.00 & 0.00 \\
            \bottomrule
        \end{tabular}
    \end{subtable}

\end{table*}
\begin{table}
\footnotesize
\centering
\ra{1.3} 
\setlength{\tabcolsep}{3pt} 
\caption{Evaluation of projector models.
Models were provided with the ground truth solutions, and were asked to evaluate them as if they were generated from evaluation subject models.
Higher is better.
Projector consistency performs nearly perfectly, far outperforming LLM-as-a-judge. 
For ``Judge w/ Answer", we additionally provided the projector model with the ground truth answer, to see if it would help improve the judge's performance.
Even in this case, the projector outperforms it. 
}
\begin{tabular}{@{}l S[table-format=2.2] S[table-format=2.2] S[table-format=2.2]@{}} 
\toprule
Projector Model & {Judge} & {Judge w/ answer} & {Projector (Ours)} \\
 \midrule
 \multicolumn{4}{@{}l}{\textbf{GPQA Diamond} - \textit{Accuracy of projector (\%)}} \\
 Llama3.1-8B & 18.69 & 64.14 & 92.93 \\
 Gemma2-9B & 57.07 & 95.96 & 97.47 \\
 Qwen2.5-7B & 44.95 & 68.18 & 98.48 \\
 Qwen2.5-14B & 38.89 & 91.92 & 98.99 \\
 Phi-4 (14B) & 79.80 & 94.95 & 98.48 \\
 Qwen2.5-32B & 62.12 & 93.94 & 99.49 \\
 \addlinespace
 \multicolumn{4}{@{}l}{\textbf{GPQA Main} - \textit{Accuracy of projector (\%)}} \\
 Llama3.1-8B & 25.00 & 65.63 & 95.76 \\
 Gemma2-9B & 60.49 & 95.31 & 95.54 \\
 Qwen2.5-7B & 45.31 & 70.09 & 97.77 \\
 Qwen2.5-14B & 39.06 & 92.41 & 99.11 \\
 Phi-4 (14B) & 83.48 & 97.32 & 98.21 \\
 Qwen2.5-32B & 65.85 & 93.08 & 99.33 \\
 \bottomrule
\end{tabular}
\label{tab:quality_of_projector}
\end{table}

\begin{table}[htbp]
\footnotesize
  \centering
  \ra{1.3} 
  \setlength{\tabcolsep}{3pt}
  \caption{Parsing failure rates (\si{\%}) for models on GPQA datasets under standard MCQA evaluation and our two stage instantiation of proposed framework. We observe a complete elimination of parsing failures when Phi-4 is chosen as projector in verifiable projection stage. LMH+ and Custom are both variants of standard MCQA evaluation with prompt engineering improvements. * - Phi-4's extreme parsing failure rate in LMH based evaluation is the result of missing chat template in LMH framework and Phi-4's high sensitivity to input format.}
  \label{tab:formatting:parsing_failure_rates}
  \sisetup{table-format=3.2} 
  \begin{tabular}{@{}lSSSS@{}}
    \toprule
    & \multicolumn{3}{c}{Standard MCQA Evaluation} & \multicolumn{1}{c}{Our Method} \\
    \cmidrule(r){2-4} \cmidrule(l){5-5}
    & {LMH} & {LMH+} & {Custom} & {Projector - Phi} \\
    \midrule
    \multicolumn{5}{@{}l}{\textbf{GPQA Diamond}} \\
    Llama3.1-8B  & 34.34  & 16.16  & 29.29 & 0.00 \\
    Gemma2-9B    & 44.95  & 19.19  &  5.56 & 0.00 \\
    Qwen2.5-7B   &  4.55  &  0.51  &  3.54 & 0.00 \\
    Qwen2.5-14B  &  0.00  &  0.00  &  0.00 & 0.00 \\
    Phi-4 (14B)  & 89.39*  &100.00*  &  0.51 & 0.00 \\
    Qwen2.5-32B  &  3.54  &  0.51  &  1.52 & 0.00 \\
    \addlinespace
    \multicolumn{5}{@{}l}{\textbf{GPQA Main}} \\
    Llama3.1-8B  & 29.02  & 13.84  & 26.79 & 0.00 \\
    Gemma2-9B    & 38.62  & 19.20  &  7.37 & 0.00 \\
    Qwen2.5-7B   &  3.13  &  0.45  &  2.01 & 0.00 \\
    Qwen2.5-14B  &  1.79  &  0.45  &  0.67 & 0.00 \\
    Phi-4 (14B)  & 91.29*  &100.00*  &  0.67 & 0.00 \\
    Qwen2.5-32B  &  2.68  &  0.67  &  0.67 & 0.00 \\
    \bottomrule
  \end{tabular}
\end{table}
\begin{table}[htbp]
\footnotesize
    \centering
    \ra{1.3} 
    \setlength{\tabcolsep}{3pt}
    \caption{Parsing failure rate for LMH, projector, and LLM-as-a-judge. Lower is better.\\ 
    * - The projector threw an error because the model's outputs could not be converted to a valid Python expression. 
    The other three evaluation methods did not throw an error because they simply marked these outputs as incorrect. 
    Llama seemed to struggle in this scenario due to its shorter context length. 
    \\ 
    $\dagger$ - For a few examples, the projector refused to choose an answer based on the model's outputs. (See section \ref{sec:parsing_failure}.)
    }
    \label{tab:judge_parsing_failure}
    \begin{tabular}{@{}l S[table-format=3.2] S[table-format=3.2] S[table-format=3.2] S[table-format=3.2]@{}} 
    \toprule

    & & & \multicolumn{2}{c}{Judge} \\ 
    \cmidrule(lr){4-5} 
    Model & {LMH} & {Projector (Phi)} & {Self} & {Phi-4} \\
    \midrule
    \multicolumn{5}{@{}l}{\textbf{ARC Challenge} - \textit{Parsing Failure (\%)}} \\
    Llama3.1-8B & NA & 0.09$^{\dagger}$ & 1.37 & 0.00\\
    Gemma2-9B & NA & 0.00 & 0.09 & 0.09\\
    Qwen2.5-7B & NA & 0.00 & 3.92 & 0.09\\
    Qwen2.5-14B & NA & 0.00 & 0.00 & 0.00\\
    Phi-4 (14B) & NA & 0.00 & 0.00 & 0.00\\
    Qwen2.5-32B & NA & 0.00 & 0.00 & 0.00\\
    \addlinespace
    \multicolumn{5}{@{}l}{\textbf{GPQA Diamond} - \textit{Parsing Failure (\%)}} \\
    Llama3.1-8B & 34.34 & 0.00 & 11.62 & 1.01\\
    Gemma2-9B & 44.95 & 0.00 & 0.00 & 0.00\\
    Qwen2.5-7B & 4.55 & 0.00 & 2.02 & 0.00\\
    Qwen2.5-14B & 0.00 & 0.00 & 1.01 & 0.00\\
    Phi-4 (14B) & 89.39 & 0.00 & 0.00 & 0.00\\
    Qwen2.5-32B & 3.54 & 0.00 & 0.51 & 0.00\\
    \addlinespace
    \multicolumn{5}{@{}l}{\textbf{GPQA Main} - \textit{Parsing Failure (\%)}} \\
    Llama3.1-8B & 29.02 & 0.00 & 7.81 & 0.00\\
    Gemma2-9B & 38.62 & 0.00 & 0.22 & 0.00\\
    Qwen2.5-7B & 3.13 & 0.00 & 2.46 & 0.00\\
    Qwen2.5-14B & 1.79 & 0.00 & 0.00 & 0.00\\
    Phi-4 (14B) & 91.29 & 0.00 & 0.00 & 0.00\\
    Qwen2.5-32B & 2.68 & 0.00 & 0.45 & 0.00\\
    \addlinespace
    \multicolumn{5}{@{}l}{\textbf{GSM8K-100} - \textit{Parsing Failure (\%)}} \\
    Llama3.1-8B & 0.00 & 10.00* & 4.00 & 1.00\\
    Gemma2-9B & 2.55 & 0.00 & 0.00 & 0.00\\
    Qwen2.5-7B & 0.00 & 0.00 & 0.00 & 0.00\\
    Qwen2.5-14B & 0.00 & 0.00 & 1.00 & 0.00\\
    Phi-4 (14B) & 0.00 & 0.00 & 0.00 & 0.00\\
    Qwen2.5-32B & 0.00 & 0.00 & 0.00 & 0.00\\
    \bottomrule
    \end{tabular}
\end{table}

\begin{table}[htbp]
\footnotesize
    \centering
    \ra{1.3} 
    \setlength{\tabcolsep}{3pt}
    \caption{Ablation study on the importance of the projector. 
    Higher is better.
    }
    \label{tab:rule_based_accuracy}
    \begin{tabular}{@{}lrrr@{}}
    \toprule
    & & \multicolumn{2}{c}{Projector} \\
    \cmidrule(lr){3-4}
    Model & LMH & Rule-based & Phi-4 \\
    \midrule
    \multicolumn{4}{@{}l}{\textbf{GPQA Diamond} - \textit{Accuracy (\%)}} \\
    Llama3.1-8B   & 18.69 & 20.71 & 32.83 \\
    Gemma2-9B     & 14.65 & 32.83 & 31.82 \\
    Qwen2.5-7B    & 29.29 & 28.79 & 34.85 \\
    Qwen2.5-14B   & 42.42 & 35.35 & 41.92 \\
    Phi-4 (14B)   & 4.04  & 34.34 & 51.52 \\
    Qwen2.5-32B   & 41.41 & 33.33 & 44.95 \\
    \addlinespace
    \multicolumn{4}{@{}l}{\textbf{GPQA Main} - \textit{Accuracy (\%)}} \\
    Llama3.1-8B   & 22.76 & 22.54 & 33.04 \\
    Gemma2-9B     & 18.30 & 33.04 & 32.59 \\
    Qwen2.5-7B    & 33.48 & 28.57 & 30.13 \\
    Qwen2.5-14B   & 38.17 & 32.37 & 37.72 \\
    Phi-4 (14B)   & 2.90  & 35.49 & 45.31 \\
    Qwen2.5-32B   & 41.74 & 34.60 & 40.85 \\
    \addlinespace
    \multicolumn{4}{@{}l}{\textbf{GSM8K-100} - \textit{Accuracy (\%)}} \\
    Llama3.1-8B   & 32.00 & 30.00 & 42.00 \\
    Gemma2-9B     & 21.00 & 59.00 & 74.00 \\
    Qwen2.5-7B    & 76.00 & 48.00 & 70.00 \\
    Qwen2.5-14B   & 80.00 & 75.00 & 84.00 \\
    Phi-4 (14B)   & 52.00 & 91.00 & 91.00 \\
    Qwen2.5-32B   & 87.00 & 95.00 & 95.00 \\
    \bottomrule
    \end{tabular}
\end{table}

\begin{table}[htbp]
\footnotesize
    \centering
    \ra{1.3} 
    \setlength{\tabcolsep}{3pt}
    \caption{Ablation study on the importance of the projector. Failure rates are reported across various models on three different benchmarks. The methods compared are a baseline (LMH) and two projector-based approaches.}
    \label{tab:rule_based_parsing_failure}
    \begin{tabular}{@{}lrrr@{}}
    \toprule
    & & \multicolumn{2}{c}{Projector} \\
    \cmidrule(lr){3-4}
    Model & LMH & Rule-based & Phi-4 \\
    \midrule
    \multicolumn{4}{@{}l}{\textbf{GPQA Diamond} - \textit{Parsing Failure (\%)}} \\
    Llama3.1-8B   & 34.34 & 38.89 & 0.00 \\
    Gemma2-9B     & 44.95 & 2.02  & 0.00 \\
    Qwen2.5-7B    & 4.55  & 13.13 & 0.00 \\
    Qwen2.5-14B   & 0.00  & 2.02  & 0.00 \\
    Phi-4 (14B)   & 89.39 & 1.52  & 0.00 \\
    Qwen2.5-32B   & 3.54  & 0.51  & 0.00 \\
    \addlinespace
    \multicolumn{4}{@{}l}{\textbf{GPQA Main} - \textit{Parsing Failure (\%)}} \\
    Llama3.1-8B   & 29.02 & 31.47 & 0.00 \\
    Gemma2-9B     & 38.62 & 1.56  & 0.00 \\
    Qwen2.5-7B    & 3.13  & 12.95 & 0.00 \\
    Qwen2.5-14B   & 1.79  & 1.12  & 0.00 \\
    Phi-4 (14B)   & 91.29 & 2.01  & 0.00 \\
    Qwen2.5-32B   & 2.68  & 1.56  & 0.00 \\
    \addlinespace
    \multicolumn{4}{@{}l}{\textbf{GSM8K-100} - \textit{Parsing Failure (\%)}} \\
    Llama3.1-8B   & 0.00  & 47.00 & 10.00 \\
    Gemma2-9B     & 2.55  & 33.00 & 0.00 \\
    Qwen2.5-7B    & 0.00  & 40.00 & 0.00 \\
    Qwen2.5-14B   & 0.00  & 16.00 & 0.00 \\
    Phi-4 (14B)   & 0.00  & 0.00  & 0.00 \\
    Qwen2.5-32B   & 0.00  & 0.00  & 0.00 \\
    \bottomrule
    \end{tabular}
\end{table}
    
\begin{table}[htbp]
\footnotesize
    \centering
    \renewcommand{\arraystretch}{1.3} 
    \setlength{\tabcolsep}{3pt}
    \caption{
    Gap in model performance depending on the choice of projector model. 
    Reported numbers are calculated by taking the difference of the max and min of the accuracies over all of the open-source projector models. 
    Judge results vary drastically depending on what model is being used as a projector. Smaller is better.} 
    \label{tab:judge_delta} 
    \begin{tabular}{@{}lrr@{}}
    \toprule
      & Projector & Judge\\
    \midrule
    \multicolumn{3}{@{}l}{\textbf{ARC Challenge} - $\Delta$ \textit{Accuracy (\%)}}\\
    Llama3.1-8B   & 0.07 & 0.18\\
    Gemma2-9B     & 0.08 & 0.32\\
    Qwen2.5-7B    & 0.07 & 0.16\\
    Qwen2.5-14B   & 0.07 & 0.12\\
    Phi-4 (14B)   & 0.07 & 0.08\\
    Qwen2.5-32B   & 0.07 & 0.10\\
    \multicolumn{3}{@{}l}{\textbf{GPQA Diamond} - $\Delta$ \textit{Accuracy (\%)}}\\
    Llama3.1-8B   & 0.04 & 0.12\\
    Gemma2-9B     & 0.07 & 0.24\\
    Qwen2.5-7B    & 0.08 & 0.12\\
    Qwen2.5-14B   & 0.10 & 0.19\\
    Phi-4 (14B)   & 0.07 & 0.15\\
    Qwen2.5-32B   & 0.08 & 0.18\\
    \multicolumn{3}{@{}l}{\textbf{GPQA Main} - $\Delta$ \textit{Accuracy (\%)}}\\
    Llama3.1-8B   & 0.05 & 0.10\\
    Gemma2-9B     & 0.04 & 0.27\\
    Qwen2.5-7B    & 0.03 & 0.13\\
    Qwen2.5-14B   & 0.04 & 0.20\\
    Phi-4 (14B)   & 0.06 & 0.17\\
    Qwen2.5-32B   & 0.05 & 0.18\\
    \bottomrule
    \end{tabular}
\end{table}

\end{document}